\documentclass[10pt,twocolumn,letterpaper]{article}

\usepackage{wacv}              
\usepackage{stfloats}  
\usepackage{afterpage}

\usepackage{amssymb}
\usepackage{graphicx}
\usepackage{svg}  
\usepackage{caption}
\usepackage{adjustbox}
\usepackage{float}

\usepackage{multirow}
\usepackage{svg}  
\usepackage{caption}
\usepackage{adjustbox}  
\definecolor{wacvblue}{rgb}{0.21,0.49,0.74}
\usepackage[pagebackref,breaklinks,colorlinks,allcolors=wacvblue]{hyperref}
\usepackage{hyperref}
\usepackage{placeins}

\def\wacvPaperID{183} 
\def\confName{WACV}
\def\confYear{2026}

\title{Surgical Gaussian Surfels: Highly Accurate Real-time Surgical Scene Rendering using Gaussian Surfels}

\author{
Idris O. Sunmola\\
Johns Hopkins University\\
Baltimore, MD, USA\\
{\tt\small isunmol1 @ jhu dot edu}
\and
Zhenjun Zhao\\
The Chinese University of Hong Kong\\
Hong Kong, China\\
{\tt\small }
\and
Samuel Schmidgall\\
Johns Hopkins University\\
Baltimore, MD, USA\\
{\tt\small }
\and
Yumeng Wang\\
Johns Hopkins University\\
Baltimore, MD, USA\\
{\tt\small }
\and
Paul Maria Scheikl\\
Johns Hopkins University\\
Baltimore, MD, USA\\
{\tt\small }
\and
Viet Pham\\
Technical University of Munich\\
Munich, Germany\\
{\tt\small }
\and
Axel Krieger\\
Johns Hopkins University\\
Baltimore, MD, USA\\
{\tt\small }
}

\begin{document}

\twocolumn[{
\renewcommand\twocolumn[1][]{#1}

\maketitle

\vspace{-0.3cm}

\begin{center}
    \includegraphics[width=\textwidth]{figs/teaser.pdf}
    \vspace{-0.3cm}
    \captionof{figure}{
    \textit{Left:} Our method achieves artifact-free renderings of surgical tool occluded areas on the StereoMIS dataset~\cite{hayoz2023stereomis}.
    \textit{Right:} 2D plot showing the fast inference speed (\textit{fps} ↑) and rendering quality (LPIPS ↓) of our method on the StereoMIS~\cite{hayoz2023stereomis} Intestine dataset.
    }
    \label{fig:teaser}
\end{center}

\vspace{0.3cm}
}]

\begin{abstract}
Accurate geometric reconstruction of deformable tissues in monocular endoscopic video remains a fundamental challenge in robot-assisted minimally invasive surgery. Although recent volumetric and point primitive methods based on neural radiance fields (NeRF) and 3D Gaussian primitives have efficiently rendered surgical scenes, they still struggle with handling artifact-free tool occlusions and preserving fine anatomical details. These limitations stem from unrestricted Gaussian scaling and insufficient surface alignment constraints during reconstruction. To address these issues, we introduce Surgical Gaussian Surfels (SGS), which transform anisotropic point primitives into surface-aligned elliptical splats by constraining the scale component of the Gaussian covariance matrix along the view-aligned axis. We also introduce the Fully Fused Deformation Multilayer Perceptron (FFD-MLP), a lightweight Multi-Layer Perceptron (MLP) that predicts accurate surfel motion fields up to 5× faster than a standard MLP. This is coupled with locality constraints to handle complex tissue deformations. We use homodirectional view-space positional gradients to capture fine image details by splitting Gaussian Surfels in over-reconstructed regions. In addition, we define surface normals as the direction of the steepest density change within each Gaussian surfel primitive, enabling accurate normal estimation without requiring monocular normal priors. We evaluate our method on two in-vivo surgical datasets, where it outperforms current state-of-the-art methods in surface geometry, normal map quality, and rendering efficiency, while remaining competitive in real-time rendering performance. We make our code available at https://github.com/aloma85/SurgicalGaussianSurfels
\end{abstract}
    
\section{INTRODUCTION}
\label{sec:intro}


Reconstructing dynamic surgical scenes from endoscopic videos is a crucial yet challenging task in robot-aided minimally invasive surgery. Accurate reconstructions not only enhance the precision of surgical instrument manipulation but also serve as a foundation for clinical applications such as AR/VR surgical simulation, medical training, surgical scene analysis, and full medical automation~\cite{dupont2021decade}. Factors such as deformable tissues, limited viewing angles, surgical tool occlusion, and severe specularity currently hinder existing 3D surface rendering methods~\cite{haidegger2019autonomy}.

The volumetric rendering method, \textit{NeRF}~\cite{mildenhall2021nerf}, was the first attempt at using a \textit{principled} ray marching approach to represent surgical scene geometry as a continuous and differentiable volume~\cite{wang2022neural,yang2023lerplane,khojasteh2025misnerf, masuda2024osnerf,ruthberg2025nerf}. However, this method has significant drawbacks, as the surgical scene is implicitly represented as neural network architecture weights and employs a probabilistic ray marching scheme to sample viewing directions, resulting in excessively slow rendering and training times~\cite{celarek2025gaussian}.

Advancements in 3D Gaussian Splatting (3DGS)~\cite{kerbl2023gaussian} have shown promise in efficiently rendering and reconstructing 3D surgical scenes. By representing scenes with explicit, topology-free Gaussian points, 3DGS facilitates rapid training and real-time rendering, outperforming neural implicit representations in both rendering speed and image quality~\cite{cheng2024gaussianpro}. Despite these advantages, 3DGS struggles with accurate geometric reconstructions, primarily due to the non-zero thickness of Gaussian point primitives, ambiguity in normal directions due to their ellipsoid-like shape, and difficulties in modeling surface edges~\cite{xie2024surface}.

To address these limitations, Gaussian Surfels have been proposed, which flatten 3D Gaussian points into 2D ellipses by setting the z-scale to zero~\cite{xie2024surface}. This transformation resolves the normal ambiguity and aligns the Gaussian points closely with the surfaces of scene objects, significantly improving optimization stability and surface reconstruction fidelity. However, applying Gaussian Surfels directly to dynamic surgical scenes presents new challenges, such as using object-centric anisotropic point primitives to model spatio-temporal deformations of soft tissues and handle surgical instrument occlusions.

In this paper, we introduce Surgical Gaussian Surfels, a novel method that combines the surface alignment capabilities of Gaussian surfels with the dynamic modeling strengths of MLPs for reconstructing deformable tissues in endoscopic videos. Our approach utilizes a fast forward mapping deformation neural network (sec.~\ref{subsec:ffd_mlp}) to model the spatiotemporal features of soft tissues, enabling flexible Gaussian motion in observation space. By defining surface normals as the direction of the steepest density change within each Gaussian primitive, we achieve accurate surface normal estimation, thereby improving reconstruction quality and addressing shape-radiance ambiguities in regions with high specular reflections.

As part of our Surgical Gaussian Surfels training pipeline, we incorporate sharp depth maps and boundary masks to enhance geometric supervision and segmentation accuracy. Sharp depths are generated by post-processing raw depth predictions with edge-aware filtering and refinement techniques (Suppl. Sec. 2), resulting in more accurate and artifact-free surface representations. Boundary masks are computed by detecting high-gradient regions in the depth maps, effectively highlighting object contours and transition zones. Integrating these sharp depths and boundary contour masks into our optimization framework provides stronger supervision at object boundaries, leading to improved rendering fidelity and faster training convergence, especially in regions with complex geometry.

To address the unique challenges of surgical environments, we utilize depth priors and a binary motion mask during Gaussian Surfel initialization (sec.~\ref{subsec:pimi}) and training. This effectively removes occluding surgical instruments from the reconstruction and reduces the motion-appearance ambiguity inherent in single-viewpoint scenarios. The efficacy of our method is shown in Fig.~\ref{fig:teaser}. Our key contributions are as follows: 

\begin{itemize}
    \item We propose Surgical Gaussian Surfels, a point-primitive-based rendering technique that combines the object surface alignment properties of surfels~\cite{pfister2000surfels} with the high-quality explicit rendering quality of 3D Gaussian Splatting to achieve high-fidelity reconstruction of dynamic surgical scenes
    \item By defining surface normals as the direction of the steepest density change within each surfel primitive, we accurately predict the orientation of surgical tissue without relying on monocular normal priors, thereby improving reconstruction geometry and mitigating issues caused by specular reflections
    \item We propose a Projection-Based Iterative Mask Integration (PIMI) method to initialize point clouds in surgical scenes, which addresses gaps and inaccuracies caused by incomplete depth data by leveraging confidence-weighted depth aggregation and refined color mapping to ensure geometrically consistent and visually accurate 3D reconstructions
    \item Based on the tiny-cuda-nn~\cite{mueller2021radiance} architecture, we develop the Fully Fused Deformation Multilayer Perceptron (FFD-MLP), a high-performance implementation of the MLP that takes advantage of highly parallelized compute architecture to predict surgical tissue deformation fields several times faster than a vanilla MLP 
    \item In quantitative and qualitative comparisons to previous attempts at dynamic surgical tissue reconstruction, our method outperforms prior implicit and explicit rendering techniques by a wide margin
\end{itemize}

\section{RELATED WORK}
\label{sec:formatting}



Recent advances in endoscopic 3D reconstruction have been driven by Neural Radiance Fields (NeRFs) and explicit representations. Early neural approaches~\cite{wang2022neural} introduced a dual network architecture that combined implicit neural fields with mask-guided ray casting for tool occlusion and depth-cueing ray marching for single-viewpoint reconstruction. Their method employs Gaussian transfer functions with stereo depth to guide point sampling near tissue surfaces. EndoSurf~\cite{wang2023endosurf} advanced this framework by representing the canonical scene through signed distance functions (SDFs), employing separate MLPs for deformation, geometry, and appearance. Their geometry network explicitly defines surfaces as zero-level sets, computing surface normals in canonical space and mapping them to observation space through deformation.

Neural LerPlane~\cite{yang2023lerplane} decomposed 4D surgical scenes by factoring them into explicit 2D planes of static and dynamic fields to address the computational overhead of volumetric neural fields. This approach treats temporal data as 4D volumes with the time axis orthogonal to 3D spatial coordinates, enabling information sharing across timesteps within static fields while reducing the negative impact of limited viewpoints. EndoGaussian~\cite{liu2024endogaussian} marked a shift toward explicit representations by introducing Holistic Gaussian Initialization (HGI) for depth-aware point cloud generation and Spatio-temporal Gaussian Tracking (SGT) for modeling surface dynamics. Their deformation field combines efficient encoding voxels with a lightweight decoder to enable real-time Gaussian tracking.

3DGS-based surgical scene reconstruction methods continue to take advantage of the rapid improvements in the explicit primitives scene reconstruction domain. EndoGS~\cite{zhu2024endogs} incorporates time-varying deformable fields by representing a surgical scene as a 4D volume. They utilize a multiresolution HexPlane~\cite{Cao2023HexPlane,kplanes_2023} to encode dynamic spatial and temporal information in the surgical scene. Deform3DGS~\cite{yang2024deform3dgs} proposes a flexible deformation modeling scheme to learn tissue deformation dynamics for each anisotropic Gaussian primitive. They improve on~\cite{Wu_2024_CVPR} by adopting learnable mean and variance operators in Fourier and polynomial basis functions to efficiently learn Gaussian point-primitive motion curves.

\begin{figure*}[t]
\centering
\includegraphics[width=\textwidth]{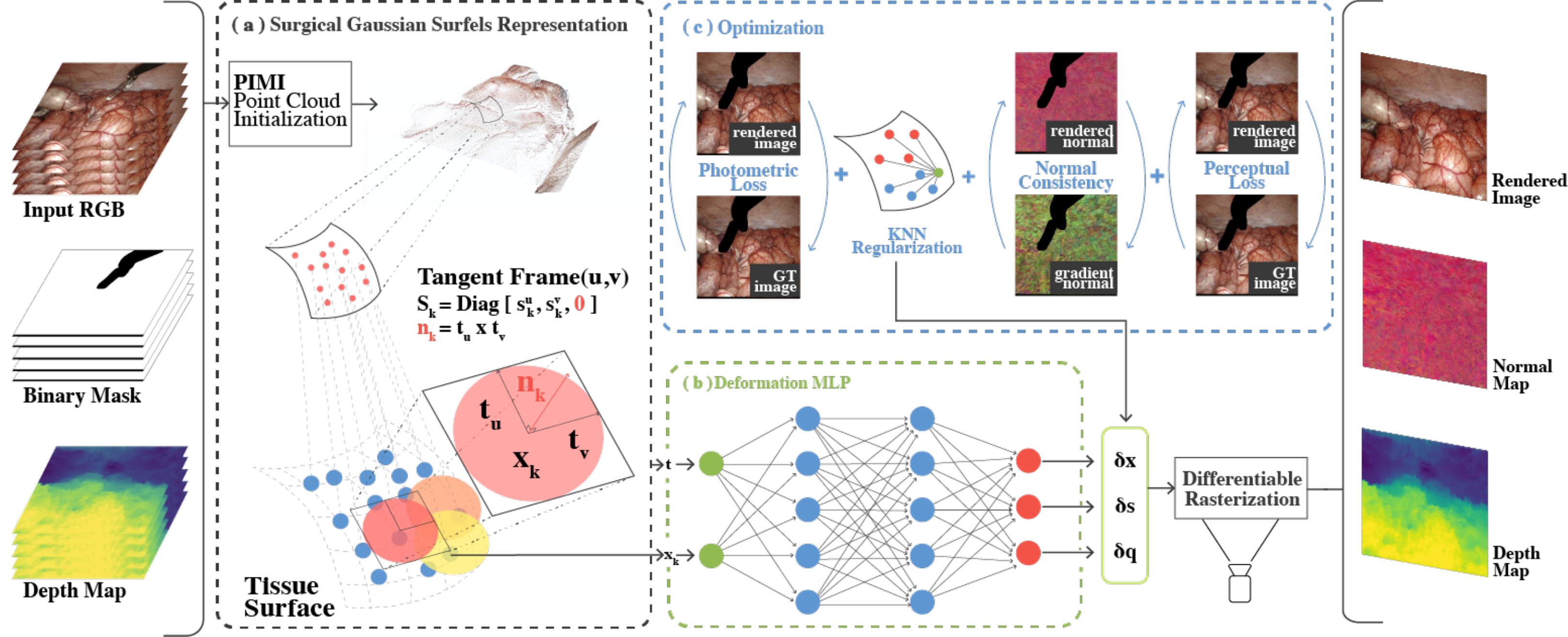}
\caption{\textbf{Method pipeline.} The full video sequence frames, Binary Mask, Depth Map is passed through PIMI (sec.~\ref{subsec:pimi}); (a) 2D points are projected to 3D then the tissue surface is represented as Gaussian Surfels with scaling matrix \(\mathbf{S}_k\), tangent vectors \(\mathbf{t}_u\) and \(\mathbf{t}_v\), and position \(\mathbf{x}_k\); (\textcolor{green}{b}) The Fully Fused Deformation prediction MLP (sec.~\ref{subsec:ffd_mlp}) takes Surfel position \(\mathbf{x}_k\) and normalized timestamp \(\mathbf{t}\) then predicts the change in position \(\delta{x}\), scale \(\delta{s}\), and rotation \(\delta{q}\); (\textcolor{blue}{c}) Optimization of Surgical Gaussian Surfels (sec.~\ref{sec:optimization}). Pipeline outputs an occlusion-free rendered surgical scene, normal map, and depth map.}
    \label{fig:architecture}
\end{figure*}

Furthermore, SurgicalGaussian~\cite{xie2024surgicalgaussian} addresses the limitation of planar structures in encoding complex motion fields using a deformation network that decouples motion and geometry. Their Gaussian initialization strategy (GIDM) uses geometry priors based on depth and mask information to reduce motion-appearance ambiguity in single-viewpoint scenarios.

Our Surgical Gaussian Surfels method builds on these advances while addressing fundamental geometric accuracy and real-time performance limitations. By constraining the z-scale of anisotropic Gaussian covariance matrices, we preserve fine geometric details that are often lost in traditional 3D Gaussian representations. Our temporal tool masking scheme with depth priors enables robust initialization in canonical space, while our lightweight MLP efficiently maps spatiotemporal deformations. Surgical tissue surface normals are represented as the direction of the steepest density change within each Gaussian primitive, thereby improving surface reconstruction quality while maintaining real-time rendering capabilities. 



\begin{figure}[t]
    \centering
    \includegraphics[width=\columnwidth]{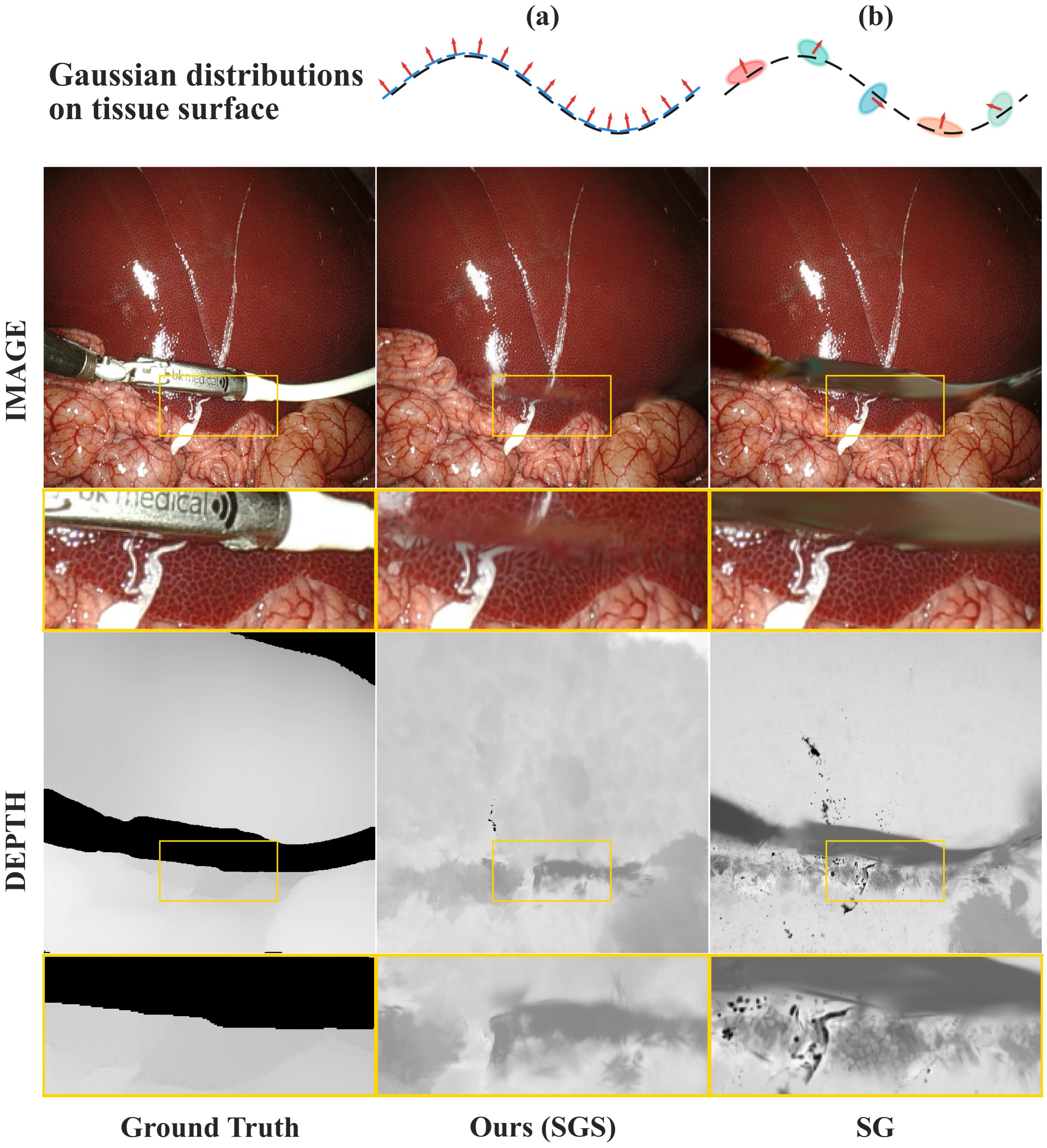}
    \caption{Qualitative comparison of (a) Our Gaussian surfel distribution-based method (SGS) with correct surface normals, Sharp Depth, and Boundary Mask infused training (Suppl. Sec. 2), (b) a Gaussian Elypsoid-based method SG~\cite{xie2024surgicalgaussian} with ambiguous surface normals.}
    \label{fig:depth_comparison}
\end{figure}

\section{METHOD}
The goal of our method is to achieve high-fidelity geometric reconstruction of deformable surgical scenes by combining the real-time rendering capabilities of 3D Gaussian Splatting with the surface-preserving properties of surfel primitives. As shown in Fig.~\ref{fig:architecture}, our pipeline takes as input a temporally aware stack of monocular endoscopic video frames, binary occlusion masks, and depth maps. The time component is normalized to $t_i = i/T$. We then initialize our Surgical Gaussian Surfels from the reconstructed point cloud, representing each point as a surface-aligned elliptical splat characterized by a constrained covariance matrix, opacity value, and view-dependent color features represented by spherical harmonic coefficients.

To model tissue deformation, we employ a fast and lightweight MLP that predicts per-surfel motion fields while maintaining temporal consistency. The network is set to minimize photometric differences between input frames and rendered views, without the need for additional monocular normal supervision. We introduce several regularization terms to enforce surface smoothness and geometric consistency in occluded regions (Sec.~\ref{sec:optimization}). Our pipeline enables real-time rendering of dynamic surgical scenes while preserving fine anatomical details and handling tool-tissue interactions.

\subsection{3D Gaussian Splatting}

3D Gaussian Splatting represents a scene as a set of anisotropic 3D Gaussian primitives \( \{ \mathcal{G}_k \mid k = 1, \dots, K \} \). Each Gaussian \( \mathcal{G}_k \) is parameterized by its position \( \mathbf{p}_k \in \mathbb{R}^3 \), opacity \( \alpha_k \in [0,1] \), rotation matrix \( \mathbf{R}_k \in \mathbb{R}^{3 \times 3} \), and scaling matrix \( \mathbf{S}_k \in \mathbb{R}^{3 \times 3} \). Together, these form a covariance matrix \( \boldsymbol{\Sigma}_k = \mathbf{O}_k \operatorname{diag}(\mathbf{s}_k) \mathbf{O}_k^T \) in world space, where its inverse \( \boldsymbol{\Sigma}_k^{-1}\) is the precision matrix~\cite{krause2025probabilisticartificialintelligence}, \( \mathbf{s}_k \in \mathbb{R}^3 \) is a scaling vector, and \( \mathbf{O}_k \in \mathbb{R}^{3 \times 3} \) is a rotation matrix parameterized by a quaternion:

\begin{equation}
\mathcal{G}_k(\mathbf{x}) = e^{-\frac{1}{2} (\mathbf{x} - \mathbf{p}_k)^T \boldsymbol{\Sigma}_k^{-1} (\mathbf{x} - \mathbf{p}_k)}
\end{equation}

For rendering, Gaussians undergo a view transformation defined by rotation \( \mathbf{R} \) and translation \( \mathbf{t} \):

\begin{equation}
\mathbf{p}_k' = \mathbf{R} \mathbf{p}_k + \mathbf{t}, \quad \boldsymbol{\Sigma}_k' = \mathbf{R} \boldsymbol{\Sigma}_k \mathbf{R}^T
\end{equation}

They are then projected to ray space through a local affine transformation:

\begin{equation}
\boldsymbol{\Sigma}_k^{2D} = \mathbf{J}_k \boldsymbol{\Sigma}_k' \mathbf{J}_k^T
\end{equation}
where \( \mathbf{J}_k \) approximates the projective transformation at the Gaussian center. View-dependent appearance is modeled using spherical harmonics, with the final color at each pixel computed through alpha composition:

\begin{equation}
c(\mathbf{x}) = \sum_{k=1}^K \alpha_k c_k \mathcal{G}_k^{2D}(\mathbf{x}) \prod_{j < k} \left(1 - \alpha_j \mathcal{G}_j^{2D}(\mathbf{x})\right)
\end{equation}

\subsection{Surgical Gaussian Surfels}
\subsubsection{Modeling}

Our Surgical Gaussian Surfels adapt the 3D Gaussian Splatting approach by introducing surface-aligned elliptical splats, optimized for rendering deformable surgical scenes. Inspired by~\cite{xie2024surface}, each surfel $\psi_k$ is parameterized by its position \(p_k \in \mathbb{R}^3\), two orthonormal tangential vectors \(t_u, t_v \in \mathbb{R}^3\), and scaling factors \(s_u, s_v \in \mathbb{R}\) that control the size of the Gaussian. The inherent primitive normal is shown in~\cref{fig:architecture}(a) as:
\begin{equation}
    t_w = t_u \times t_v,
\end{equation}
representing the direction of the steepest change in surfel density. This formulation enables direct gradient-based optimization, ensuring accurate alignment of surfels with the surface of the surgical tissue.

The position of any point on the 2D Gaussian in world space is given by:
\begin{equation}
    P(u, v) = p_k + s_u t_u u + s_v t_v v, \quad u, v \in \mathbb{R},
\end{equation}
where \((u, v)\) are local coordinates in the tangent plane. Using homogeneous coordinates, the transformation of the Gaussian from local to world space is represented as:
\begin{equation}
    H =
    \begin{bmatrix}
        s_u t_u & s_v t_v & 0 & p_k \\
        0 & 0 & 0 & 1
    \end{bmatrix},
\end{equation}
where \(H \in \mathbb{R}^{4 \times 4}\) encodes scaling, rotation, the zeroed \(\mathbf{z}\)-component, and translation. As such, \( \mathbf{R}_k \operatorname{diag}(\mathbf{s}_k) = H_k[:,:3] \). It then follows that the density of each Surgical Gaussian Surfel primitive is modeled as:
\begin{equation}
    \mathcal{G}_k(u, v) = \exp\left(-\frac{u^2 + v^2}{2}\right).
\end{equation}

\subsection{PIMI Gaussian Surfel Primitive Initialization}
\label{subsec:pimi}
Xie et al.~\cite{xie2024surgicalgaussian} initialize a surgical scene point cloud by projecting visible pixels from depth maps and binary masks into 3D space. However, when depth data are incomplete, this results in noticeable gaps in the reconstructed point cloud and inaccuracies in the initialization due to color and geometric incongruencies.

To mitigate these issues, we propose an initialization method, termed \textit{Projection-Based Iterative Mask Integration (PIMI)} (Fig.~\ref{fig:architecture}(a)), which improves the quality and completeness of the initial point cloud by leveraging confidence-weighted depth aggregation and aggregated color mapping. The depth map $D_t$ for each frame is pre-processed to create a confidence-weighted depth map $D^*$ using a thresholded confidence mask, which ensures the exclusion of noisy depth data. Simultaneously, we aggregate the color maps $C_t$ to compute the refined color map $C^*$. This aggregation scheme at timestep $t$ is defined as:
\begin{equation}
    D^* = \frac{\sum_{t=1}^T D_t \cdot W_t}{\sum_{t=1}^T W_t}, \quad C^* = \frac{\sum_{t=1}^T C_t \cdot W_t}{\sum_{t=1}^T W_t}
\end{equation}
where $W_t$ is the confidence mask derived by identifying reliable depth values within the 2nd to 99th percentile range of the depth distribution. Using $D^*$ and $C^*$, we generate the refined point cloud $P^*$. Each 2D pixel coordinate $(x, y)$ is projected into 3D space based on the intrinsic camera parameters $K$ as:
\begin{equation}
    p(x, y) = D^*(x, y) \cdot K^{-1} \begin{bmatrix} x \\ y \\ 1 \end{bmatrix}, \quad
    P^* = \{(p_i, C_i^*)\}_{i=1}^N,
    \label{eq:3d_projection_point_cloud}
\end{equation}
where $P^*$ is the final point cloud, with $p_i$ as the 3D points and $C_i^*$ as their RGB colors encoded via spherical harmonics coefficients. This sequential integration of depth and color ensures geometric consistency and accurate color representation of the projected surgical scene.

\begin{figure}[t]
    \centering
    \includegraphics[width=\columnwidth]{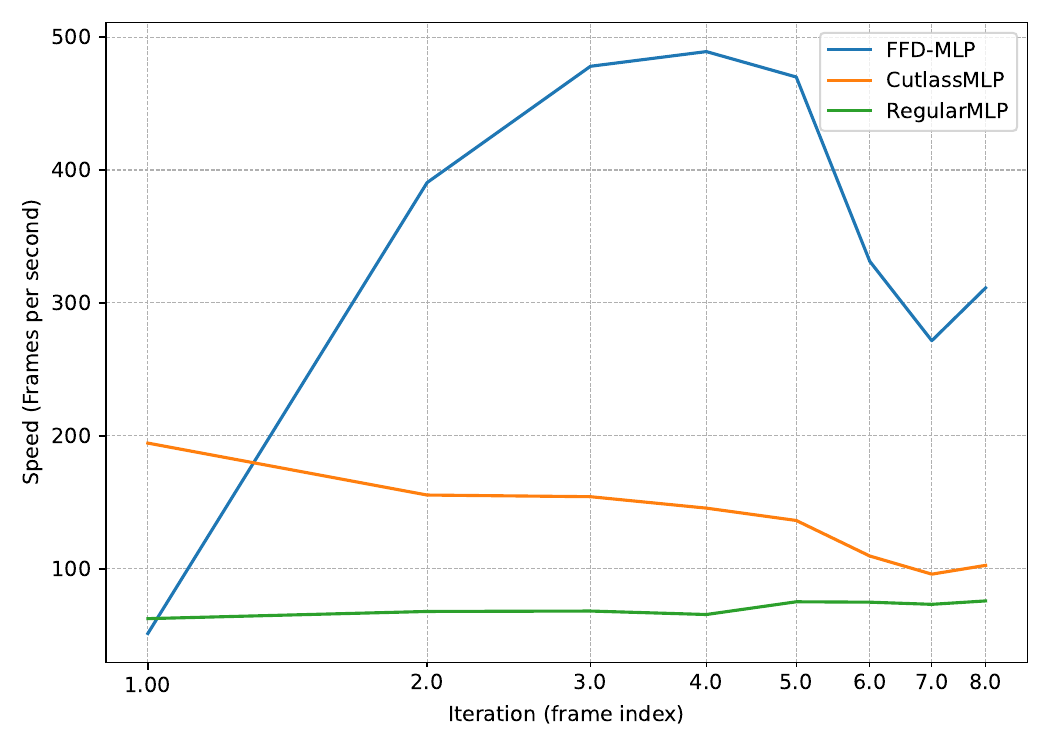}
    \caption{Quantitative comparison of FFD-MLP, CutlassMLP~\cite{nvidia2021cutlass}, and Vanilla MLP on the EndoNerf~\cite{wang2022neural} pulling dataset. Our implementation for deformable tissue prediction is, at times, 5× faster than other MLP implementations at inference time.}
    \label{fig:mlp_graph}
\end{figure}

\subsection{Fully Fused Deformation MLP (FFD-MLP)}
\label{subsec:ffd_mlp}
For our tissue deformation network in Fig.~\ref{fig:architecture}\hyperref[fig:arch-b]{(b)}, we aim to develop an architecture that is both fast and accurate in predicting motion fields. Inspired by~\cite{sun20243dgstream} and~\cite{M_ller_2022}, we devise a fully fused implementation of the MLP on modern GPUs to predict surgical scene deformation. We refer to our network as the Fully Fused Deformation MLP (FFD-MLP), a high-performance implementation of the multilayer perceptron designed explicitly for fast deformation network training within the SGS pipeline. 

Built on Tiny-CUDA-NN~\cite{mueller2021radiance}, FFD-MLP achieves a significant speedup through GPU-optimized kernels that fuse multiple operations into a single CUDA kernel, eliminating memory bandwidth bottlenecks and maximizing computational throughput. The network architecture consists of 11 fully connected hidden layers, each with 128 neurons, processing 72-dimensional encoded features (63 spatial and 9 temporal) to produce a 10-dimensional deformation output (3, 4, and 3 for the position, rotation, and scaling of each Gaussian surfel, respectively). 

The network employs a composite frequency encoding that matches the positional encoding used in standard NeRF-style~\cite{mildenhall2021nerf} networks: $\gamma(\mathbf{p}, t) = [\gamma_{xyz}(\mathbf{p}), \gamma_t(t)]$, where $\mathbf{p}\in\mathbb{R}^3$ represents 3D coordinates and $t \in \mathbb{R}$ represents time. The spatial encoding $\gamma_{xyz}$ uses 10 frequency bands: $\gamma_{xyz}(\mathbf{p}) = [\mathbf{p}, \sin(2^0\pi\mathbf{p}), \cos(2^0\pi\mathbf{p}), \ldots, \sin(2^9\pi\mathbf{p}), \cos(2^9\pi\mathbf{p})]$, while the temporal encoding $\gamma_t$ uses 4 frequency bands: $\gamma_t(t) = [t, \sin(2^0\pi t), \cos(2^0\pi t), \ldots, \sin(2^3\pi t), \cos(2^3\pi t)]$. This encoding provides multi-scale frequency information, enabling the network to learn both fine-grained local deformations and coarse global transformations. The core innovation lies in the fully fused CUDA implementation, which leverages NVIDIA's Tensor Cores through Warp Matrix Multiply Accumulate (WMMA)~\cite{nvidia2017tensorcores} operations, processing 16×16×16 matrix multiplications in a single instruction while storing activations in shared memory and using register-based storage for weight matrices.





Empirical results demonstrate significant performance improvements—2.5$\times$ faster training, 5$\times$ higher iteration throughput (Fig.~\ref{fig:mlp_graph}), and a 40\% reduction in GPU memory—while maintaining superior rendering quality. Given
\begin{equation}
    (\delta\mathbf{x}, \delta\mathbf{s}, \delta\mathbf{q}) = \text{FFD-MLP}\!\left(\gamma(\mathbf{p}, t)\right),
\end{equation}
The predicted translation $\delta\mathbf{x}$, scale $\delta\mathbf{s}$, and rotation $\delta\mathbf{q}$ update the canonical parameters of each surfel:
\begin{equation}
\mathbf{p}' = \mathbf{p} + \delta\mathbf{x},\quad
\mathbf{s}' = \mathbf{s} \odot \delta\mathbf{s},\quad
\mathbf{q}' = \mathbf{q} \otimes \delta\mathbf{q}.
\end{equation}
where $\odot$ denotes element‑wise scaling and $\otimes$ quaternion composition.

\subsection{Optimization}
\label{sec:optimization}

To capture fine surgical tissue detail and preserve geometric surfel--surface alignment, we optimize a weighted objective (see Fig.~\ref{fig:architecture}(c)) combining photometric fidelity, structural similarity, spatial smoothing, deformation regularization, perceptual alignment (DINO~\cite{caron2021emerging}), depth supervision, and normal consistency. The total objective is

\begin{multline}
    L_{\text{total}} = L_{\text{photo}} +
    \lambda_{\text{smooth}} L_{\text{TV}} + \lambda_{\text{pos}} L_{\text{pos}} + \lambda_{\text{cov}} L_{\text{cov}} + \\[-2pt]
    \lambda_{\text{per}} L_{\text{per}} + \lambda_{\text{normal}} L_{\text{n}} + \lambda_{\text{depth}} L_{\text{depth}}.
    \label{eq:total_loss}
\end{multline}

\paragraph{Image Reconstruction (Photometric + SSIM + TV).}
Pixel‑wise fidelity is enforced with an $L_1$ + SSIM blend. We penalize discrepancies in pixel intensity while accounting for surgical tool-masked regions:
\begin{multline}
    L_{\text{photo}} = (1 - \lambda_{\text{SSIM}}) \| I - \hat{I} \|_1 \\[-2pt]
    {}+ \lambda_{\text{SSIM}} \bigl(1 - \text{SSIM}(I, \hat{I})\bigr),
\end{multline}
where \(I\) and \(\hat{I}\) are ground‑truth and rendered images. Spatial pixel smoothness is encouraged with total variation:
\begin{equation}
L_{\text{TV}} = \sum_{p,q} \Bigl( \| \hat{I}_{p,q} - \hat{I}_{p-1,q} \|_1 + \| \hat{I}_{p,q} - \hat{I}_{p,q-1} \|_1 \Bigr).
\end{equation}

\paragraph{Deformation Regularization (Position + Scale).}
We encourage locally coherent motion and scale evolution across neighboring surfels $\mathcal{N}(i)$:
\begin{equation}
L_{\text{pos}} = \sum_{i} \sum_{j \in \mathcal{N}(i)} \bigl\| x_i - x_j \bigr\|_1 - \bigl\| \hat{x}_i - \hat{x}_j \bigr\|_1 ,
\end{equation}
\begin{equation}
L_{\text{cov}} = \sum_{i} \sum_{j \in \mathcal{N}(i)} \bigl\| s_i - s_j \bigr\|_1 - \bigl\| \hat{s}_i - \hat{s}_j \bigr\|_1 ,
\end{equation}
where \(x_i\) and \(x_j\) are the positions of neighboring surfels in the initial state, and \(\hat{x}_i\) and \(\hat{x}_j\) are their positions after deformation. In addition, \(s_i\) and \(s_j\) represent the scaling factors of neighboring surfels.

\paragraph{Perceptual (DINO) Loss.}
High‑level appearance is matched using ViT‑based DINO features:
\begin{equation}
L_{\text{per}} = 1 - \cos\!\bigl(\phi(\hat{I}), \phi(I)\bigr),
\end{equation}
where $\phi$ represents the DINO~\cite{caron2021emerging} multiscale semantic embedding extractor.

\paragraph{Depth Supervision.}
Rendered depth is nudged toward sharp, mask‑filtered supervision using:
\begin{equation}
L_{\text{depth}} = \| D_{\text{prediction}} - D_{\text{GT}} \|_1 .
\end{equation}

\paragraph{Normal Consistency.}
We encourage agreement between rendered normals and expected geometric normals from depth signals via a two‑target cosine ranking:
\begin{multline}
L_{\text{n}} = \alpha \bigl(1 - \cos(\mathbf{n}_{\text{rendered}}, \mathbf{n}_{\text{median}})\bigr) \\[-2pt]
{}+ \beta \bigl(1 - \cos(\mathbf{n}_{\text{rendered}}, \mathbf{n}_{\text{expected}})\bigr),
\end{multline}
where $\mathbf{n}_{\text{median}}$ and $\mathbf{n}_{\text{expected}}$ are derived from scene geometry along the viewing ray; $\alpha=0.6$, $\beta=0.4$ in all experiments.


\section{Experiments}
\begin{table*}[t]
\caption{Quantitative comparisons of our method SGS with EndoNeRF~\cite{wang2022neural}, EndoSurf~\cite{wang2023endosurf}, LerPlane~\cite{yang2023lerplane}, EndoGaussian~\cite{liu2024endogaussian}, SurgicalGaussian (SG)~\cite{xie2024surgicalgaussian}, EndoGS~\cite{zhu2024endogs}, and Deform3DGS~\cite{yang2024deform3dgs}.}
\centering
\begin{tabular}{c|c|ccc|ccc|cc}
\hline
\multirow{2}{*}{Dataset} & \multirow{2}{*}{Methods} & \multicolumn{3}{c|}{"pulling"} & \multicolumn{3}{c|}{"cutting"} & \multicolumn{2}{c}{Rendering} \\
\cline{3-10}
& & LPIPS↓ & PSNR↑ & SSIM↑ & LPIPS↓ & PSNR↑ & SSIM↑ & FPS↑ & GPU↓ \\
\hline\hline
\multirow{5}{*}{EndoNeRF} & EndoNeRF & 0.080 & 29.088 & 0.930 & 0.106 & 26.792 & 0.901 & 0.04 & 15 GB \\
& EndoSurf & 0.111 & 36.919 & 0.961 & 0.103 & 35.551 & 0.955 & 0.05 & 17 GB \\
& LerPlane & 0.085 & 36.272 & 0.936 & 0.114 & 33.774 & 0.901 & 1.02 & 20 GB \\
& EndoGaussian & 0.089 & 36.429 & 0.951 & 0.102 & 35.773 & 0.956 & 190 & \textbf{2} GB \\
& SG & 0.049 & 38.783 & 0.970 & 0.062 & 37.505 & 0.961 & 80 & 4 GB \\
& EndoGS & 0.053 & 37.240 & 0.961 & 0.043 & 35.696 & 0.955 & 55 & 3 GB \\
& Deform3DGS & 0.063 & 38.367 & 0.961 & 0.042 & \textbf{38.771} & 0.956 & 225 &  5 GB\\
& \textbf{Ours} & \textbf{0.027} & \textbf{39.059} & \textbf{0.971} & \textbf{0.027} & 37.609 & \textbf{0.965} & \textbf{310} & \textbf{2} GB \\

\hline\hline
\multirow{2}{*}{} & \multirow{2}{*}{} & \multicolumn{3}{c|}{"intestine"} & \multicolumn{3}{c|}{"liver"} & FPS↑ & GPU↓ \\
\cline{3-10}
\multirow{5}{*}{StereoMIS} & EndoNeRF & 0.153 & 30.510 & 0.833 & 0.316 & 27.370 & 0.680 & 0.06 & 13 GB \\
& EndoSurf & 0.204 & 29.660 & 0.853 & 0.248 & 28.941 & 0.820 & 0.08 & 14 GB \\
& LerPlane & 0.206 & 29.441 & 0.822 & 0.254 & 28.852 & 0.793 & 1.45 & 19 GB \\
& EndoGaussian & 0.213 & 29.024 & 0.805 & 0.295 & 26.174 & 0.728 & 200 & \textbf{2} GB \\
& SG & 0.145 & 31.496 & 0.890 & 0.135 & 31.668 & 0.893 & 140 & 3 GB \\
& EndoGS & 0.060 & \textbf{35.239} & 0.934 & 0.051 & \textbf{34.540} & 0.932 & 61 & 2 GB \\
& Deform3DGS & 0.131 & 32.589 & 0.897 & 0.110 & 32.929 & 0.900 & 196 & 5 GB \\
& \textbf{Ours} & \textbf{0.048} & 34.669 & \textbf{0.936} & \textbf{0.047} & 34.390 & \textbf{0.934} & \textbf{423} & \textbf{2} GB \\
\hline
\end{tabular}
\label{tab:comparison}
\end{table*}
\subsection{Experiment settings}
\subsubsection{Datasets and evaluation.}
We evaluate our method on two publicly available datasets, EndoNerf~\cite{wang2022neural} and StereoMIS~\cite{hayoz2023stereomis}. The EndoNerf dataset captures two cases of an in-vivo da Vinci robot prostatectomy from a single panoptic viewpoint. Each video comprises complex surgical tissue deformation and surgical tool occlusion. We also evaluate on two  $\sim$5\textit{s} video segments from the videos P2\_7 and P2\_8 of the StereoMIS dataset, which account for the breathing of porcine subjects and tissue deformation during a liver and intestinal procedure. Both datasets are sampled at a frame rate of 30 \textit{fps}. We follow~\cite{wang2023endosurf} and divide the training and test sets in a 7:1 ratio.  We estimate scene depth on the StereoMIS dataset using a pre-trained Omnidata~\cite{eftekhar2021omnidata} model and surgical tool occlusion binary masks generated by a pre-trained SAM 2~\cite{ravi2024sam2} model. We measure the surgical scene reconstruction capabilities of SGS using three standard metrics: LPIPS, PSNR, and SSIM. We achieve state-of-the-art (SOTA) results on all metrics. We clearly show our qualitative and quantitative results against prior works in Fig.~\ref{fig:qualitative_comparison} and Table~\ref{tab:comparison}, respectively.
\subsubsection{Implementation Details.}
All experiments are carried out on an NVIDIA RTX A6000 using the PyTorch framework. Adam~\cite{kingma2014adam} optimizer is used during training. The loss weights in Eq.~\ref{eq:total_loss} are set to:
\begin{multline}
\lambda_{SSIM} = 0.2, \quad \lambda_{smooth} = 0.006, \quad \lambda_{pos} = 1.0, \\ \quad \lambda_{cov} = 200.0, \quad \lambda_{per} = 1.0, \quad \lambda_{depth} = 0.0001.
\end{multline}

\afterpage{%
\begin{figure*}[h]
    \centering
    \includegraphics[width=\linewidth]{figs/compare.pdf} 
    \caption{Qualitative comparisons on the EndoNeRF dataset~\cite{wang2022neural} of our method (SGS) with EndoNeRF~\cite{wang2022neural}, EndoSurf~\cite{wang2023endosurf}, LerPlane~\cite{yang2023lerplane}, EndoGaussian~\cite{liu2024endogaussian}, SurgicalGaussian~\cite{xie2024surgicalgaussian}, EndoGS~\cite{zhu2024endogs}, and Deform3DGS~\cite{yang2024deform3dgs}.}
    \label{fig:qualitative_comparison}
\end{figure*}
}

\subsection{Main Results}
We compare SGS with existing methods for real-time rendering of surgical scenes in the EndoNeRF~\cite{wang2022neural} and StereoMIS~\cite{hayoz2023stereomis} datasets. SDF-based approaches such as EndoSurf~\cite{wang2023endosurf} require separate MLPs for deformation, geometry, and appearance, which increases computational complexity and limits real-time performance. Neural LerPlane~\cite{yang2023lerplane} factorizes 4D scenes into 2D planes to reduce overhead, but struggles to capture complex deformations. EndoGaussian~\cite{liu2024endogaussian} shifts towards explicit depth-aware point clouds but relies heavily on pre-initialized geometry, making it less adaptable to dynamic changes.

Furthermore, EndoGS~\cite{huang2024endodgs} represents a surgical scene as a 4D volume, leveraging a multi-resolution HexPlane~\cite{Cao2023HexPlane} to encode both spatial and temporal dynamics. Deform3DGS~\cite{yang2024deform3dgs} enhances tissue deformation modeling by incorporating learnable mean and variance operators in Fourier and polynomial basis functions. Both methods face challenges relating to the complexity of representing highly dynamic and non-rigid tissue deformations, as well as the computational cost of handling multi-resolution representations and learnable basis functions, which may limit real-time applicability and efficiency in practical surgical scenarios.

Lastly, SurgicalGaussian~\cite{xie2024surgicalgaussian} introduces a deformation network that decouples motion and geometry, but struggles to render fine details in the surgical scene. SGS mostly outperforms these previous methods in all datasets, achieving the lowest LPIPS, the highest PSNR, and the highest SSIM, while qualitatively capturing specularity and anatomical features more accurately, as shown in Fig.~\ref{fig:qualitative_comparison} and Table~\ref{tab:comparison}, respectively.


\subsubsection{Ablation Study}
\begin{table}[t]
\centering
\caption{Ablation study on EndoNeRF~\cite{wang2022neural} dataset.}
\label{tab:ablation}
\begin{adjustbox}{max width=\columnwidth}
\begin{tabular}{lcccccc}
\hline
& \multicolumn{3}{c}{"pulling"} & \multicolumn{3}{c}{"cutting"} \\ 
\cline{2-7}
Model & LPIPS$\downarrow$ & PSNR$\uparrow$ & SSIM$\uparrow$ & LPIPS$\downarrow$ & PSNR$\uparrow$ & SSIM$\uparrow$ \\ 
\hline\hline
w/o Perceptual loss & 0.028 & 38.911 & 0.970 & 0.032 & 37.307 & 0.962 \\ 
w/o Photometric loss & 0.028 & 38.852 & 0.970 & 0.043 & 35.566 & 0.957 \\ 
w GIDM initialization & \textbf{0.027} & 38.907 & 0.970 & 0.031 & 37.380 & 0.963 \\ 
Full model (PIMI init) & \textbf{0.027} & \textbf{39.059} & \textbf{0.971} & \textbf{0.027} & \textbf{37.609} & \textbf{0.965} \\ 
\hline
\end{tabular}
\end{adjustbox}
\end{table}

We perform ablation studies in a "leave-one-out" fashion on the EndoNerf \cite{wang2023endosurf} dataset. To assess the impact of our PIMI Gaussian initialization scheme, we replace it with the GIDM \cite{xie2024surgicalgaussian} point cloud initialization. We also provide test-time results without perceptual, photometric, and TV loss. The results are summarized in Table \ref{tab:ablation}. The similarity in the ablation results to the full model could be attributed to the homodirectional view-space positional gradient guidance \cite{xie2024absgs} during Gaussian surfel densification at training time. We also conduct an ablation study on the surgical scene deformation prediction network (Fig.~\ref{fig:architecture}(b)) within our rendering pipeline. We replace our FFD-MLP (sec.~\ref{subsec:ffd_mlp}) with the CutlassMLP~\cite{nvidia2021cutlass} during both training and inference times, along with a separate training and inference run with a "vanilla" MLP implementation. As shown in Fig.~\ref{fig:mlp_graph}, our FFD-MLP attains inference speeds 5× faster than both the CutlassMLP and classical MLP implementations.

\subsubsection{Comparison on EndoNerf Dataset}
The EndoNerf dataset~\cite{wang2022neural} is a mainstay in the surgical scene reconstruction domain. The dataset comprises six curated clips from DaVinci robotic prostatectomy procedures. Each clip spans 4$\sim$8 seconds and is captured from a fixed stereo camera viewpoint. The dataset features complex surgical scenes with non-rigid tissue deformation and frequent tool occlusion. We qualitatively and quantitatively compare our method using the EndoNerf dataset against SOTA methods~\cite{wang2022neural,wang2023endosurf,yang2023lerplane,liu2024endogaussian,xie2024surgicalgaussian,zhu2024endogs,yang2024deform3dgs} and highlight our results in Fig.~\ref{fig:qualitative_comparison} and Table~\ref{tab:comparison}, respectively. We train and test our method on tissue pulling and tissue cutting videos, where it achieves lower perceptual error (LPIPS), higher signal fidelity (PSNR), and stronger structural consistency (SSIM). This indicates that SGS renders dynamic tissue motion fields closer to the ground-truth surgical tissue motion and photometric features at the pixel level. As a result of our efficiency gains from FFD-MLP (sec.~\ref{subsec:ffd_mlp}), this improvement in rendering quality does not come at the expense of rendering efficiency and minimal GPU memory usage, demonstrating the practical deployability of SGS in resource-constrained surgical robotics settings.

\subsubsection{Comparison on StereoMIS Dataset}
The StereoMIS dataset~\cite{hayoz2023stereomis} comprises 11 in‑vivo porcine sequences captured at 15fps by the DaVinci Xi stereo endoscope. The dataset captures complex physiological motion of surgical tissue undergoing breathing-induced deformation, instrument occlusion, and manipulation. We benchmark SGS against state-of-the-art methods~\cite{wang2022neural,wang2023endosurf,yang2023lerplane,liu2024endogaussian,xie2024surgicalgaussian,zhu2024endogs,yang2024deform3dgs} in representative intestine and liver sequences (Table~\ref{tab:comparison}), achieving the lowest perceptual error (LPIPS) and the highest structural similarity (SSIM) in both scenes, demonstrating that SGS more accurately reconstructs fine-grained tissue appearance and geometry. EndoGS~\cite{zhu2024endogs}, which models the scene as a time-varying 4D volume via multi-resolution HexPlane~\cite{Cao2023HexPlane} encoding, yields marginally higher PSNR in the liver case, indicating its strength in signal fidelity when leveraging high-capacity spatiotemporal representations. Crucially, SGS matches or exceeds the overall reconstruction quality of EndoGS while delivering higher frame rates and maintaining a smaller GPU footprint.

\section{Conclusion}
In this article, we introduce Surgical Gaussian Surfels (SGS), an explicit rendering method that takes advantage of the accurate geometry reconstruction capabilities of surface-aligned anisotropic splats. We utilize scale-constrained elliptical primitives to produce high-fidelity renders of surgical scenes. We define surface normals as the direction of the steepest density change, thereby generating accurate surgical tissue surface normals without normal priors. In addition, we introduce FFD-MLP, an implementation of the tiny-cuda-nn for extremely fast surgical tissue deformation field prediction. Furthermore, by leveraging both Projection-Based Iterative Mask Integration (PIMI) for robust point-cloud initialization and Gaussian surfel densification based on homodirectional view-space positional gradients, our method achieves state-of-the-art reconstruction quality with real-time rendering performance. A limitation of our current implementation is that training, although on the order of minutes, cannot be done on the fly; a common setback of volumetric rendering and point primitive methods. SGS has the potential to advance surgical scene understanding and support a range of intraoperative applications, including surgical navigation and robot-aided surgical procedures.
{
    \small
    \bibliographystyle{ieeenat_fullname}
    \bibliography{main}

\begin{thebibliography}{34}
\providecommand{\natexlab}[1]{#1}
\providecommand{\url}[1]{\texttt{#1}}
\expandafter\ifx\csname urlstyle\endcsname\relax
  \providecommand{\doi}[1]{doi: #1}\else
  \providecommand{\doi}{doi: \begingroup \urlstyle{rm}\Url}\fi

\bibitem[Cao and Johnson(2023)]{Cao2023HexPlane}
Ang Cao and Justin Johnson.
\newblock Hexplane: A fast representation for dynamic scenes.
\newblock \emph{CVPR}, 2023.

\bibitem[Caron et~al.(2021)Caron, Touvron, Misra, J\'egou, Mairal, Bojanowski,
  and Joulin]{caron2021emerging}
Mathilde Caron, Hugo Touvron, Ishan Misra, Herv\'e J\'egou, Julien Mairal,
  Piotr Bojanowski, and Armand Joulin.
\newblock Emerging properties in self-supervised vision transformers.
\newblock In \emph{Proceedings of the International Conference on Computer
  Vision (ICCV)}, 2021.

\bibitem[Celarek et~al.(2025)Celarek, Kopanas, Drettakis, Wimmer, and
  Kerbl]{celarek2025gaussian}
A. Celarek, G. Kopanas, G. Drettakis, M. Wimmer, and B. Kerbl.
\newblock Does 3d gaussian splatting need accurate volumetric rendering?
\newblock \emph{Computer Graphics Forum}, 2025.
\newblock Article e70032.

\bibitem[Cheng et~al.(2024)Cheng, Long, Yang, Yao, Yin, Ma, Wang, and
  Chen]{cheng2024gaussianpro}
K. Cheng, X. Long, K. Yang, Y. Yao, W. Yin, Y. Ma, W. Wang, and X. Chen.
\newblock Gaussianpro: 3d gaussian splatting with progressive propagation.
\newblock In \emph{Forty-first International Conference on Machine Learning},
  2024.

\bibitem[Dai et~al.(2024)Dai, Xu, Xie, Liu, Wang, and Xu]{xie2024surface}
P. Dai, J. Xu, W. Xie, X. Liu, H. Wang, and W. Xu.
\newblock High-quality surface reconstruction using gaussian surfels.
\newblock In \emph{ACM SIGGRAPH 2024 Conference Papers}, pages 1--11, 2024.

\bibitem[Dupont et~al.(2021)Dupont, Nelson, Goldfarb, Hannaford, Menciassi,
  O'Malley, Simaan, Valdastri, and Yang]{dupont2021decade}
P.~E. Dupont, B.~J. Nelson, M. Goldfarb, B. Hannaford, A. Menciassi, M.~K.
  O'Malley, N. Simaan, P. Valdastri, and G.-Z. Yang.
\newblock A decade retrospective of medical robotics research from 2010 to
  2020.
\newblock \emph{Science Robotics}, 6\penalty0 (60), 2021.

\bibitem[Eftekhar et~al.(2021)Eftekhar, Sax, Malik, and
  Zamir]{eftekhar2021omnidata}
A. Eftekhar, A. Sax, J. Malik, and A. Zamir.
\newblock Omnidata: A scalable pipeline for making multi-task mid-level vision
  datasets from 3d scans.
\newblock In \emph{Proceedings of the IEEE/CVF International Conference on
  Computer Vision}, pages 10786--10796, 2021.

\bibitem[Haidegger(2019)]{haidegger2019autonomy}
T. Haidegger.
\newblock Autonomy for surgical robots: Concepts and paradigms.
\newblock \emph{IEEE Transactions on Medical Robotics and Bionics}, 1\penalty0
  (2):\penalty0 65--76, 2019.

\bibitem[Hayoz and Max(2023)]{hayoz2023stereomis}
M. Hayoz and A. Max.
\newblock Stereomis.
\newblock Zenodo, 2023.
\newblock \url{https://doi.org/10.5281/zenodo.7727692}.

\bibitem[Huang et~al.(2024)Huang, Cui, Bai, Guo, Xu, and Ren]{huang2024endodgs}
Y. Huang, B. Cui, L. Bai, Z. Guo, M. Xu, and H. Ren.
\newblock Endo-4dgs: Distilling depth ranking for endoscopic monocular scene
  reconstruction with 4d gaussian splatting.
\newblock \emph{arXiv preprint arXiv:2401.16416}, 2024.

\bibitem[Kerbl et~al.(2023)Kerbl, Kopanas, Leimk{\"u}hler, and
  Drettakis]{kerbl2023gaussian}
B. Kerbl, G. Kopanas, T. Leimk{\"u}hler, and G. Drettakis.
\newblock 3d gaussian splatting for real-time radiance field rendering.
\newblock \emph{ACM Transactions on Graphics}, 42\penalty0 (4):\penalty0
  139--1, 2023.

\bibitem[Khojasteh et~al.(2025)Khojasteh, Fuentes-Jimenez, Pizarro, Espinel,
  and Bartoli]{khojasteh2025misnerf}
S.B. Khojasteh, D. Fuentes-Jimenez, D. Pizarro, Y. Espinel, and A. Bartoli.
\newblock {MIS-NeRF}: Neural radiance fields in minimally-invasive surgery.
\newblock \emph{International Journal of Computer Assisted Radiology and
  Surgery}, 20\penalty0 (7):\penalty0 1481--1490, 2025.

\bibitem[Kingma and Ba(2014)]{kingma2014adam}
D.P. Kingma and J. Ba.
\newblock Adam: A method for stochastic optimization.
\newblock \emph{arXiv preprint arXiv:1412.6980}, 2014.

\bibitem[Krause and
  Hübotter(2025)]{krause2025probabilisticartificialintelligence}
Andreas Krause and Jonas Hübotter.
\newblock Probabilistic artificial intelligence, 2025.

\bibitem[Liu et~al.(2024)Liu, Li, Yang, and Yuan]{liu2024endogaussian}
Y. Liu, C. Li, C. Yang, and Y. Yuan.
\newblock Endogaussian: Gaussian splatting for deformable surgical scene
  reconstruction.
\newblock \emph{arXiv preprint arXiv:2401.12561}, 2024.

\bibitem[Masuda et~al.(2024)Masuda, Hachiuma, Saito, Kajita, and
  Takatsume]{masuda2024osnerf}
M. Masuda, R. Hachiuma, H. Saito, H. Kajita, and Y. Takatsume.
\newblock {OS-NeRF}: Generalizable novel view synthesis for occluded
  open-surgical scenes.
\newblock In \emph{2024 IEEE Conference on Virtual Reality and 3D User
  Interfaces Abstracts and Workshops (VRW)}, pages 345--350, Orlando, FL, USA,
  2024. IEEE.

\bibitem[Mildenhall et~al.(2021)Mildenhall, Srinivasan, Tancik, Barron,
  Ramamoorthi, and Ng]{mildenhall2021nerf}
B. Mildenhall, P.~P. Srinivasan, M. Tancik, J.~T. Barron, R. Ramamoorthi, and
  R. Ng.
\newblock {NeRF}: Representing scenes as neural radiance fields for view
  synthesis.
\newblock \emph{Communications of the ACM}, 65\penalty0 (1):\penalty0 99--106,
  2021.

\bibitem[M{\"u}ller et~al.(2021)M{\"u}ller, Rousselle, Nov{\'a}k, and
  Keller]{mueller2021radiance}
Thomas M{\"u}ller, Fabrice Rousselle, Jan Nov{\'a}k, and Alexander Keller.
\newblock Real-time neural radiance caching for path tracing.
\newblock \emph{ACM Trans. Graph.}, 40\penalty0 (4), 2021.

\bibitem[Müller et~al.(2022)Müller, Evans, Schied, and Keller]{M_ller_2022}
Thomas Müller, Alex Evans, Christoph Schied, and Alexander Keller.
\newblock Instant neural graphics primitives with a multiresolution hash
  encoding.
\newblock \emph{ACM Transactions on Graphics}, 41\penalty0 (4):\penalty0
  1–15, 2022.

\bibitem[{NVIDIA Developer}(2017)]{nvidia2017tensorcores}
{NVIDIA Developer}.
\newblock Programming tensor cores in cuda 9.
\newblock
  \url{https://developer.nvidia.com/blog/programming-tensor-cores-cuda-9/},
  2017.
\newblock Accessed: 2025-07-16.

\bibitem[{NVIDIA Developer}(2021)]{nvidia2021cutlass}
{NVIDIA Developer}.
\newblock Cutlass: Fast linear algebra in cuda c++.
\newblock \url{https://developer.nvidia.com/blog/cutlass-linear-algebra-cuda/},
  2021.
\newblock Accessed: 2025-07-16.

\bibitem[Pfister et~al.(2000)Pfister, Zwicker, van Baar, and
  Gross]{pfister2000surfels}
H. Pfister, M. Zwicker, J. van Baar, and M. Gross.
\newblock Surfels: Surface elements as rendering primitives.
\newblock In \emph{Proceedings of the 27th Annual Conference on Computer
  Graphics and Interactive Techniques (SIGGRAPH '00)}, pages 335--342, 2000.

\bibitem[Ravi et~al.(2024)Ravi, Gabeur, Hu, Hu, Ryali, Ma, and
  Khedr]{ravi2024sam2}
N. Ravi, V. Gabeur, Y.-T. Hu, R. Hu, C. Ryali, T. Ma, and H.~et~al. Khedr.
\newblock {SAM 2}: Segment anything in images and videos.
\newblock \emph{arXiv preprint arXiv:2408.00714}, 2024.

\bibitem[Ruthberg et~al.(2025)Ruthberg, Bly, Gunderson, Chen, Alighezi, Seibel,
  and Abuzeid]{ruthberg2025nerf}
J.S. Ruthberg, R. Bly, N. Gunderson, P. Chen, M. Alighezi, E.J. Seibel, and
  W.M. Abuzeid.
\newblock Neural radiance fields (nerf) for 3d reconstruction of monocular
  endoscopic video in sinus surgery.
\newblock \emph{Otolaryngology--Head and Neck Surgery}, 172\penalty0
  (4):\penalty0 1435--1441, 2025.
\newblock Epub 2025 Jan 10.

\bibitem[{Sara Fridovich-Keil and Giacomo Meanti} et~al.(2023){Sara
  Fridovich-Keil and Giacomo Meanti}, Warburg, Recht, and
  Kanazawa]{kplanes_2023}
{Sara Fridovich-Keil and Giacomo Meanti}, Frederik~Rahbæk Warburg, Benjamin
  Recht, and Angjoo Kanazawa.
\newblock K-planes: Explicit radiance fields in space, time, and appearance.
\newblock In \emph{CVPR}, 2023.

\bibitem[Sun et~al.(2024)Sun, Jiao, Li, Zhang, Zhao, and
  Xing]{sun20243dgstream}
Jiakai Sun, Han Jiao, Guangyuan Li, Zhanjie Zhang, Lei Zhao, and Wei Xing.
\newblock 3dgstream: On-the-fly training of 3d gaussians for efficient
  streaming of photo-realistic free-viewpoint videos.
\newblock In \emph{Proceedings of the IEEE/CVF Conference on Computer Vision
  and Pattern Recognition (CVPR)}, pages 20675--20685, 2024.

\bibitem[Wang et~al.(2022)Wang, Long, Fan, and Dou]{wang2022neural}
Y. Wang, Y. Long, S.~H. Fan, and Q. Dou.
\newblock Neural rendering for stereo 3d reconstruction of deformable tissues
  in robotic surgery.
\newblock In \emph{International Conference on Medical Image Computing and
  Computer-Assisted Intervention}, pages 431--441, Cham, 2022. Springer Nature
  Switzerland.

\bibitem[Wu et~al.(2024)Wu, Yi, Fang, Xie, Zhang, Wei, Liu, Tian, and
  Wang]{Wu_2024_CVPR}
Guanjun Wu, Taoran Yi, Jiemin Fang, Lingxi Xie, Xiaopeng Zhang, Wei Wei, Wenyu
  Liu, Qi Tian, and Xinggang Wang.
\newblock 4d gaussian splatting for real-time dynamic scene rendering.
\newblock In \emph{Proceedings of the IEEE/CVF Conference on Computer Vision
  and Pattern Recognition (CVPR)}, pages 20310--20320, 2024.

\bibitem[Xie et~al.(2024)Xie, Yao, Cao, Lin, Tang, Dong, and
  Guo]{xie2024surgicalgaussian}
W. Xie, J. Yao, X. Cao, Q. Lin, Z. Tang, X. Dong, and X. Guo.
\newblock Surgicalgaussian: Deformable 3d gaussians for high-fidelity surgical
  scene reconstruction.
\newblock In \emph{International Conference on Medical Image Computing and
  Computer-Assisted Intervention}, pages 617--627, Cham, 2024. Springer Nature
  Switzerland.

\bibitem[Yang et~al.(2023)Yang, Wang, Wang, Yang, and Shen]{yang2023lerplane}
C. Yang, K. Wang, Y. Wang, X. Yang, and W. Shen.
\newblock Neural lerplane representations for fast 4d reconstruction of
  deformable tissues.
\newblock In \emph{International Conference on Medical Image Computing and
  Computer-Assisted Intervention}, pages 46--56, Cham, 2023. Springer Nature
  Switzerland.

\bibitem[Yang et~al.(2024)Yang, Li, Shen, Gong, Dou, and
  Jin]{yang2024deform3dgs}
Shuojue Yang, Qian Li, Daiyun Shen, Bingchen Gong, Qi Dou, and Yueming Jin.
\newblock Deform3dgs: Flexible deformation for fast surgical scene
  reconstruction with gaussian splatting, 2024.

\bibitem[Ye et~al.(2024)Ye, Li, Liu, Qiao, and Dou]{xie2024absgs}
Z. Ye, W. Li, S. Liu, P. Qiao, and Y. Dou.
\newblock Absgs: Recovering fine details in 3d gaussian splatting.
\newblock In \emph{Proceedings of the 32nd ACM International Conference on
  Multimedia}, pages 1053--1061, 2024.

\bibitem[Zha et~al.(2023)Zha, Cheng, Li, Harandi, and Ge]{wang2023endosurf}
R. Zha, X. Cheng, H. Li, M. Harandi, and Z. Ge.
\newblock Endosurf: Neural surface reconstruction of deformable tissues with
  stereo endoscope videos.
\newblock In \emph{International Conference on Medical Image Computing and
  Computer-Assisted Intervention}, pages 13--23, Cham, 2023. Springer Nature
  Switzerland.

\bibitem[Zhu et~al.(2024)Zhu, Wang, Cui, Jin, Lin, and Yu]{zhu2024endogs}
Lingting Zhu, Zhao Wang, Jiahao Cui, Zhenchao Jin, Guying Lin, and Lequan Yu.
\newblock Endogs: Deformable endoscopic tissues reconstruction with gaussian
  splatting, 2024.

\end{thebibliography}
}

\end{document}